# Stemming Hallucination in Language Models Using a Licensing Oracle

*Richard Ackermann[1], Simeon Emanuilov[2]*

[1]*RA Software, San Diego, United States*, [2]*Department of Software Technologies, Faculty of Mathematics and Informatics, Sofia University "St. Kliment Ohridski", Bulgaria*
Emails: *richard@rasoftware.co, ssemanuilo@fmi.uni-sofia.bg*

*Language models exhibit remarkable natural language generation capabilities but remain prone to hallucinations, generating factually incorrect information despite producing syntactically coherent responses. This study introduces the Licensing Oracle, an architectural solution designed to stem hallucinations in LMs by enforcing truth constraints through formal validation against structured knowledge graphs. Unlike statistical approaches that rely on data scaling or fine-tuning, the Licensing Oracle embeds a deterministic validation step into the model's generative process, ensuring that only factually accurate claims are made. We evaluated the effectiveness of the Licensing Oracle through experiments comparing it with several state-of-the-art methods, including baseline language model generation, fine-tuning for factual recall, fine-tuning for abstention behavior, and retrieval-augmented generation (RAG). Our results demonstrate that although RAG and fine-tuning improve performance, they fail to eliminate hallucinations. In contrast, the Licensing Oracle achieved perfect abstention precision (AP = 1.0) and zero false answers (FAR-NE = 0.0), ensuring that only valid claims were generated with 89.1% accuracy in factual responses. This work shows that architectural innovations, such as the Licensing Oracle, offer a necessary and sufficient solution for hallucinations in domains with structured knowledge representations, offering guarantees that statistical methods cannot match. Although the Licensing Oracle is specifically designed to address hallucinations in fact-based domains, its framework lays the groundwork for truth-constrained generation in future AI systems, providing a new path toward reliable, epistemically grounded models.*

## 1. Introduction

Language Models (LMs)[1] have demonstrated impressive capabilities in natural language generation; however, they remain fundamentally unreliable in terms of factual accuracy, often producing statements that are syntactically correct but empirically false. This phenomenon, commonly referred to as "hallucination," arises from the limitations of the transformer architecture, which is optimized for generating fluent sequences of tokens based on statistical patterns rather than grounding output in verifiable knowledge (Kalai et al., 2025).

In our prior study, "How Large Language Models Are Designed to Hallucinate" (Ackermann & Emanuilov, 2025) we argued that hallucination is not merely an artifact of insufficient training data or poor fine-tuning but rather a structural feature of the transformer model itself. Specifically, transformers function as coherence engines that lack the mechanisms to regulate epistemic commitments or enforce truth constraints. Consequently, LLMs generate plausible-sounding claims without being able to distinguish between true and false information.

---

[1] We use the terms "language models" (LMs) and "large language models" (LLMs) interchangeably throughout this work, as the distinction has become increasingly fluid across models of varying scales.

To address this fundamental issue, we propose the concept of a Licensing Oracle, an architectural enhancement that integrates the formal verification of factual claims into the generative process. Unlike statistical learning methods, which rely on data patterns and probabilistic reasoning, the Licensing Oracle enforces epistemic boundaries through deterministic validation against a structured knowledge graph and the The Shapes Constraint Language (SHACL) rules. This ensures that LMs can generate claims that are both logically consistent and supported by verifiable evidence, thereby eliminating hallucinations regarding structured factual knowledge.

In this study, we build upon the theoretical foundation established in our earlier work and present the results of a series of experiments designed to evaluate the effectiveness of the Licensing Oracle. Our experiments involved five distinct approaches to factual question-answering: baseline LM generation, fine-tuning for factual recall, fine-tuning for abstention, retrieval-augmented generation, and graph-based RAG that incorporates the Licensing Oracle for formal validation. The experimental results demonstrate that while statistical methods (e.g., fine-tuning and RAG) can influence factual accuracy, they cannot eliminate hallucinations in a reliable or deterministic manner. In contrast, the Licensing Oracle achieves near-perfect abstention precision (AP = 1.0) and effectively eliminates false answers, providing a structural solution to the problem of hallucination that statistical learning alone cannot solve (see **Table 5**).

The contribution of this work lies in demonstrating that hallucinations about structured knowledge, such as factual claims about entities and relationships that can be formally represented in a knowledge graph, can be mitigated through architectural enforcement rather than through better data or larger models. By embedding formal validation into the generative process, we provide an architectural solution that scales across domains and is generalizable beyond factual question answering tasks. This study shows that the Licensing Oracle offers deterministic guarantees for factual accuracy, which purely statistical approaches fail to provide.

## 2. Related Work

The challenge of LLM hallucination has spawned diverse research efforts in statistical learning, architectural design, and formal verification. This section situates our Licensing Oracle within this landscape, examining how prior studies have approached the hallucination problem and identifying the specific gaps that our architecture addresses. We organize our review around three key questions: (1) Can statistical methods alone solve hallucinations? (2) What architectural solutions exist, and what are their limitations? (3) How can formal validation be integrated into generative AI systems?

Our analysis reveals a critical pattern: while statistical approaches improve performance metrics, they cannot provide deterministic guarantees; existing architectural solutions, such as RAG ground generation, lack epistemic discipline; and formal validation techniques remain disconnected from the generative process. The Licensing Oracle bridges these gaps by architecturally enforcing truth constraints through real-time validation, moving beyond retrieval-augmented generation to constraint-governed generations.

### 2.1 The Nature and Origins of Hallucination

The academic community remains divided on the fundamental nature of LLM hallucinations. One perspective views hallucinations as correctable artifacts of training data quality, biased corpora, and suboptimal inference procedures (Ji et al., 2023). Kalai et al., 2025 argue that LLMs are trained with evaluation incentives that "reward guessing over acknowledging uncertainty," suggesting that refined training objectives could mitigate this behavior. This statistical perspective implies that hallucinations represent a solvable engineering challenge.

In contrast, the architectural perspective, which underpins this study, posits that hallucinations are an inevitable consequence of the transformer architecture itself (Xu et al., 2024). From this perspective, LMs function as "coherence engines" optimized for statistical plausibility rather than factual veracity, lacking structural mechanisms to ground generated claims in verifiable evidence (Ackermann & Emanuilov, 2025).

Xu et al., 2024 formally demonstrate that hallucination is inevitable because LLMs, as computable functions, cannot provably learn all other computable functions representing ground truth (Xu et al., 2025).

In our prior work (Ackermann & Emanuilov, 2025), we developed this architectural thesis through an ontological analysis of the self-attention mechanism of the transformer. We argue that transformers simulate relational structures of meaning, tokens acquiring significance through reference to other tokens, without the existential grounding in temporality, mood, and care that stabilizes human understanding. This creates what we term a "flat semantic space,"[2] where continuations are governed by statistical coherence rather than world-disclosure. We distinguished between ontological hallucinations, which arise when a flat semantic space violates existential structures such as temporality or affordances, and residual reasoning hallucinations, in which models mimic inference by recycling linguistic traces of human reasoning. This theoretical foundation motivates our current work, which postulates that if hallucinations are structural rather than statistical, structural interventions are required.

## 2.2 Statistical Mitigation Approaches and Their Limitations

### 2.2.1 Fine-Tuning for Factual Grounding

Supervised fine-tuning of domain-specific factual data represents the most direct statistical approach to improving model reliability. However, this method faces documented limitations, including catastrophic forgetting of pre-trained knowledge (Kirkpatrick et al., 2017; McCloskey & Cohen, 1989) and "ripple effect failures" where edits to one fact adversely affect unrelated knowledge (J. Wu et al., 2025; Zhang et al., 2024). Most critically, Gekhman et al., 2024 demonstrated a "fine-tuning paradox": training LLMs on new factual knowledge can linearly increase their propensity to hallucinate, as models learn to generate confident but incorrect responses when uncertain. Our experimental results replicate this finding, with fine-tuning degrading accuracy from 16.7% to 8.5%, providing strong empirical support for the inadequacy of parameter optimization (**Table 2**).

### 2.2.2 Training for Epistemic Abstention

An alternative statistical approach trains models to abstain from answering when uncertain, commonly termed "epistemic abstention" (Wen et al., 2025). Methods include inferred abstention, which quantifies uncertainty through internal model states such as semantic entropy (Kuhn et al., 2023), and supervised abstention, exemplified by R-Tuning (Zhang et al., 2024), which explicitly fine-tunes models to output "I don't know" for questions outside their parametric knowledge. However, these approaches achieve only probabilistic and unreliable abstention behaviors (C.-K. Wu et al., 2025). Our Gemma-Abstain model, which implements the R-Tuning methodology, achieved only 56.7% abstention precision—barely better than random chance—demonstrating that models learn linguistic patterns rather than genuine epistemic boundaries (see **Table 2**). This failure of learned abstention motivates our deterministic architectural alternative.

## 2.3 Retrieval-Augmented Generation: Capabilities and Gaps

Retrieval-Augmented Generation, represents the dominant architectural solution for grounding LMs. RAG combines parametric memory (model weights) with non-parametric memory (external knowledge bases), retrieving relevant documents at inference time to provide verifiable context. Our baseline experiments confirm RAG's effectiveness, improving accuracy from 50.1% to 89.5% (see **Table 3**). However, RAG suffers from well-documented limitations including retrieval of irrelevant or contradictory information and inability to synthesize conflicting sources (Gao et al., 2024; H. Wang et al., 2025; Y. Wang, Ren, et al., 2024).

---

[2] We define "flat semantic space" as the condition where tokens derive meaning solely through reference to other tokens, lacking grounding in world-disclosure or temporality (Ackermann & Emanuilov, 2025).

Most critically, standard RAG lacks mechanisms for principled abstention, a gap termed "epistemic mismatch" by Ghafouri et al., 2025. This mismatch describes models expressing unwarranted certainty even when retrieved information is insufficient or contradictory (Sun et al., 2025). Magesh et al., 2025 demonstrated that commercial RAG systems in legal research still hallucinate in 17-33% of cases, contradicting industry claims of "hallucination-free" systems. This epistemic gap—the inability to distinguish lack of evidence from evidence of absence—represents the specific architectural problem our Licensing Oracle addresses.

## 2.4 Graph-Based Retrieval and Formal Validation

Recent work has explored replacing unstructured text corpora with structured Knowledge Graphs (KGs) for retrieval. Graph-RAG proponents argue that KGs, which store information as explicit triples (subject, predicate, object), reduce semantic ambiguity and better support multi-hop reasoning. However, comparative studies yield mixed results. (Olausson et al., 2023) found that vector-RAG and Graph-RAG have distinct, complementary strengths suited to different query types. Our results (89.5% for embedding-RAG vs. 89.1% for Graph-RAG) support the task-equivalence view for retrieval performance (see **Table 3** and **Table 4**).

However, we argue that the primary value of KGs lies not in retrieval performance but in their formal verifiability (Dehal et al., 2025; Khan et al., 2025). A structured RDF graph can be deterministically validated through symbolic query languages, while high-dimensional vector spaces cannot. This verifiability is the essential prerequisite for our Licensing Oracle architecture.

Parallel work in neuro-symbolic AI has demonstrated the value of decoupling probabilistic generation from deterministic reasoning (Liang et al., 2025; Qi et al., 2025). Olausson et al., 2023 introduced LINC, which uses LMs as semantic parsers to translate natural language into First-Order Logic, then executes formal theorem provers for deterministic validation. Similarly, constrained decoding methods integrate formal grammars directly into token-level generation to guarantee syntactic correctness (Geng et al., 2024; Tuccio et al., 2025). Our work extends this paradigm from syntactic to semantic constraints, using SHACL validation as a mandatory licensing gate for factual claims (Ma et al., 2025; Y. Wang, Chen, et al., 2024).

## 2.5 SHACL and Formal Constraint Validation

The Shapes Constraint Language, a W3C standard for validating RDF knowledge graphs, enables formal specification of structural and logical constraints. Recent work has explored SHACL in LLM contexts, primarily using LLMs to generate SHACL constraints from natural language Westermann et al., 2025 or to interpret validation reports. Our architecture is novel in its control flow: rather than using LLMs to write SHACL, we use SHACL to govern LLM generation in real-time, implementing a mandatory validation gate that ensures epistemic grounding before emission (Publio & Labra Gayo, 2025; Tian et al., 2025).

## 2.6 Research Gap and Contribution

The literature demonstrates that (1) statistical approaches alone cannot eliminate hallucinations, with fine-tuning often exacerbating the problem (Gekhman et al., 2024); (2) standard RAG lacks deterministic abstention mechanisms, suffering from epistemic mismatch; and (3) the components for a solution, Graph-RAG, neuro-symbolic validation, and SHACL constraints, exist but remain unintegrated.

Our contribution synthesizes these components into a novel architecture that architecturally couples LLMs with formal knowledge graphs, using SHACL validation as a real-time licensing oracle. Building on the theoretical foundation established in our prior work (Ackermann & Emanuilov, 2025), which identified hallucination as a structural consequence of flat semantic space, this paper provides the empirical validation and engineering implementation of a solution. We move beyond "retrieval-augmented generation" to "constraint-governed generation," achieving deterministic, provable epistemic abstention that directly addresses the epistemic reliability gaps left open by all prior approaches (see **Figure 3** for a capability comparison).

Orthogonal to our architectural approach, alignment methods like Constitutional AI (Bai et al., 2022) and RLHF attempt to reduce hallucinations through training objectives that reward truthfulness. While these statistical methods show promise, they cannot provide the deterministic guarantees achieved through formal validation. Similarly, knowledge editing methods like ROME (Meng et al., 2023) enable targeted fact updates but struggle with scaling and consistency. Our Licensing Oracle is complementary to these approaches, providing architectural enforcement that could be combined with improved training objectives.

## 3. Theoretical Framework and Design of the Licensing Oracle

At the core of our approach lies the Licensing Oracle, a formal architectural component designed to govern the generative process of LMs. The Licensing Oracle operates as a control layer that enforces truth constraints during text generation by validating each factual claim against a structured knowledge graph and a set of logical constraints encoded using SHACL. This mechanism prevents the generation of hallucinated statements by rejecting unverified or logically inconsistent claims before they are emitted by the model. The Licensing Oracle addresses the "flat semantic space" problem identified in our prior work (Ackermann & Emanuilov, 2025) by introducing ontological grounding through formal validation. Where transformers lack existential anchoring to world-disclosure, the Oracle enforces being-constrained-by relations to verifiable entities in a knowledge graph. This architectural intervention provides the structural grounding that attention mechanisms alone cannot achieve.

The Licensing Oracle intervenes directly within the model's generative process. It introduces a deterministic validation step that explicitly ensures that each claim made by the model is supported by verifiable evidence grounded in a knowledge base that is formal, consistent, and logically constrained.

This architectural addition shifts the focus from data scaling and model size to epistemic grounding. The model's output is no longer governed solely by probabilistic language patterns but by structural mechanisms that demand factual accuracy and logical consistency.

### 3.1 The Knowledge Graph and SHACL Constraints

The knowledge graph is central to the Licensing Oracle's operation. For this work, we employ an RDF-based graph populated with structured knowledge from sources such as DBpedia, enhanced with domain-specific ontologies and formal relationships[3]. The graph is designed to capture structured knowledge, such as factual claims about entities (e.g., rivers and philosophers) and their relationships (e.g., "flows through," "life span").

The graph serves as both a source of knowledge and a validator. When the LLM generates a factual claim, it is first extracted as a triple (subject, predicate, object). This triple is then validated against the knowledge graph, which functions as a licensing oracle. Using SHACL constraints[4], the graph enforces domain-specific rules, such as:

- **Type constraints**: Ensuring that a river flows through a state, not the other way around.

- **Logical constraints**: Validating that the elevation of a river's source is greater than its mouth for gravity-fed flow.

- **Temporal constraints**: Ensuring that relationships, such as philosophical influence, are logically consistent within time constraints (e.g., lifespan overlap).

---

[3] DBpedia SPARQL endpoint: https://dbpedia.org/sparql, Structured knowledge extracted from Wikipedia
[4] W3C SHACL Specification: https://www.w3.org/TR/shacl/, Shapes Constraint Language for RDF validation

If a claim passes the validation step, it is licensed for emission; otherwise, the system outputs an abstention token, preventing the generation of false or unverified statements.

## 3.2 Model Integration and Validation Process

The integration of the Licensing Oracle into an LLM is achieved through a middleware layer that operates alongside the model's standard generation pipeline. During inference, as the model generates text in streaming mode, the following process occurs:

1. **Claim Extraction:** The generated text is parsed to extract potential factual claims using a Named Entity Recognition (NER) model like GLiNER (Zaratiana et al., 2025)[5], which identifies entities and maps verbs to predicates, creating triples (subject, predicate, object)[6].

2. **SHACL Validation:** The extracted triples are validated against the knowledge graph using the SHACL rules, ensuring the claims are factually consistent and logically valid.

3. **Licensing Decision:** If the claim passes the validation, it is licensed and allowed to proceed as part of the model's output. If the claim fails validation, the system outputs an abstain response, ensuring no unverified or incorrect claim is generated.

4. **Provenance Reporting**: For transparency and interpretability, the system logs which facts or constraints supported the decision, providing full provenance of the generated content. This makes the model's reasoning process more transparent and traceable.

This layered approach ensures that the model's outputs are not only syntactically coherent but also epistemically grounded. The Licensing Oracle serves as an additional safeguard, ensuring that the LLM adheres to a higher standard of factual reliability by enforcing structural constraints on its generation process.

## 3.3 Comparison to Other Hallucination Mitigation Methods

While methods like fine-tuning and RAG improve model performance by augmenting training data or providing contextual support, they do not address the fundamental epistemic gap in LLMs, namely their inability to verify the truthfulness of the claims they generate.

- **Fine-Tuning:** While fine-tuning LLMs on factual data improves recall, it often results in catastrophic forgetting or overfitting to surface patterns. Our experiments show that fine-tuned models fail to achieve reliable abstention behavior, with only 56.7% precision in abstaining from false claims, which is slightly better than random.

- **RAG:** RAG improves performance by retrieving relevant contextual information, but it remains probabilistic, it retrieves and generates based on the statistical relevance of the context, not factual correctness. As a result, it still generates hallucinations when the retrieved context is inaccurate or incomplete.

In contrast, the Licensing Oracle provides a deterministic solution to hallucinations by validating each generated claim against formal knowledge representations. It ensures that the model only generates claims that are epistemically valid, eliminating false assertions through abstention.

## 4. Experimental Setup

The primary objective of the experiments conducted in this work is to empirically validate the effectiveness of the Licensing Oracle in mitigating hallucinations in language models. Specifically, we aim to demonstrate that the architectural coupling of LMs with a formal verification system provides a deterministic solution to hallucinations about structured knowledge, hallucinations that arise from the model's inability to validate factual claims against verifiable knowledge.

---

[5] GLiNER uses bidirectional transformers for zero-shot entity recognition without task-specific fine-tuning
[6] We note that our evaluation assumes perfect claim extraction. In practice, NER errors could cause false abstentions (failing to extract valid claims) or false validations (extracting incorrect triples that happen to match the graph). Measuring extraction accuracy independently is an important direction for future work.

To achieve this, we compare the Licensing Oracle against several other hallucination mitigation methods, including baseline LM generation, fine-tuning for factual recall, fine-tuning for abstention behavior, and retrieval-augmented generation. We also evaluate the Licensing Oracle in conjunction with graph-based RAG, which integrates retrieval with relational validation.

## 4.1 Dataset Construction

The experimental evaluations were conducted using a dataset of question-answer pairs derived from structured knowledge about U.S. rivers. The dataset was specifically designed to test factual grounding and epistemic discipline. A total of 17,726 question-answer pairs were generated, targeting entities (e.g., river names, geographical features) and relationships (e.g., tributary connections, river lengths, and discharge rates) that are formalized in a knowledge graph.

## 4.2 Data Acquisition and Augmentation

The core data was extracted from DBpedia's SPARQL endpoint[7], targeting 9,538 U.S. river entities. The dataset includes detailed attributes such as hydrological metrics (length, discharge), geographical relationships (source and mouth locations), and administrative data (state, county). Due to the inherent sparsity of some attributes in DBpedia, we employed an LM-based deterministic augmentation pipeline to infer missing data such as alternative river names and missing hydrological measurements. This created a more robust dataset for evaluating factual accuracy in a broader context. We acknowledge that using LLM-generated data to augment the validation graph introduces a potential circularity concern. Approximately 15% of graph attributes were LLM-augmented, primarily for missing alternative names and select hydrological measurements. Future work should validate the Oracle exclusively against human-curated or authoritative sources.

## 4.3 Experimental Conditions

We evaluated five distinct approaches to factual question answering, representing the current state-of-the-art and new innovations. The conditions were as follows:

- **Baseline LLM Evaluation:** Three pre-trained language models, Claude Sonnet 4.5, Google Gemini 2.5 Flash Lite, and Google Gemma 3-4B-Instruct, were evaluated on the question-answer dataset without any grounding mechanisms.

- **Fine-Tuning for Factual Recall:** We fine-tuned the Google Gemma 3-4B-Instruct model on the entire dataset, with a focus on factual accuracy. This condition serves as a baseline for performance improvements via parameter optimization.

- **Fine-Tuning for Abstention Behavior:** A second fine-tuning regime was applied to Google Gemma 3-4B-Instruct, this time focusing on training the model to abstain from generating answers when uncertain. This was achieved by replacing incorrect answers in the training data with an "I don't know" response.

- **Retrieval-Augmented Generation:** The RAG approach integrates embedding-based retrieval with language model generation. Relevant documents are retrieved using semantic embeddings and injected into the model's context window. This method provides additional context to improve factual accuracy without explicitly validating the generated output.

---

[7] SPARQL Protocol: https://www.w3.org/TR/sparql11-protocol/ - Standard query language for RDF graphs

- **Graph-Based RAG with Licensing Oracle:** This method builds upon RAG by incorporating a Licensing Oracle, which validates the claims generated by the LLM against a structured knowledge graph. Here, the knowledge graph is not just used for retrieval but serves as a licensing gate, ensuring that all generated claims are consistent with formal domain knowledge and SHACL constraints.

## 4.4 Evaluation Metrics

To evaluate the effectiveness of each method, we introduced a set of metrics designed to measure both factual accuracy and epistemic discipline:

### 4.4.1. Accuracy

The percentage of correct answers among all questions attempted. This measures the model's ability to generate factually accurate responses.

$$\text{Accuracy} = \frac{\text{Correct Answers}}{\text{Total Questions}} \times 100\,\%$$

**Equation 1.** Accuracy metric measuring the percentage of factually correct answers among all questions attempted, without accounting for abstention behavior.

### 4.4.2. Abstention Precision (AP)

Among all abstentions, the percentage that were appropriate, i.e., the claim should not have been generated because there was insufficient evidence or it violated logical constraints. This metric is crucial for assessing the effectiveness of abstention mechanisms.

$$AP = \frac{\text{Appropriate Abstentions}}{\text{Total Abstentions}} = \frac{|\{q \in Q : \text{abstain}(q) \land \neg \text{entailed}(q, \mathcal{G})\}|}{|\{q \in Q : \text{abstain}(q)\}|}$$

$$Where: (Q) = questions, (\mathcal{G}) = knowledge\,graph$$

**Equation 2.** Abstention Precision quantifies the fraction of model abstentions that were epistemically appropriate, i.e., cases where the claim was not entailed by the knowledge graph $\mathcal{G}$.

### 4.4.3. Constraint Violation Rejection Rate (CVRR)

The fraction of contradictory claims (those that violate SHACL constraints) that are correctly rejected by the system. This metric is particularly important for evaluating the formal validation capabilities of the Licensing Oracle.

$$CVRR = \frac{|\{(s,p,o) : \text{violates}((s,p,o), \Phi) \land \text{rejected}((s,p,o))\}|}{|\{(s,p,o) : \text{violates}((s,p,o), \Phi)\}|}$$

$$Where: ((s,p,o)) = triple, (\Phi) = SHACL\,constraints$$

**Equation 3.** Constraint Violation Rejection Rate measures the system's ability to detect and reject triples (s, p, o) that violate formal SHACL constraints Φ, indicating the effectiveness of logical consistency enforcement.

4.4.4. False Answer Rate on Non-Entailed Claims (FAR-NE)

The fraction of claims that were generated when they should have been abstained (i.e., the claim was neither supported by the knowledge graph nor logically valid). This measures the frequency of hallucinated responses despite the system's abstention capabilities.

$$\text{FAR-NE} = \frac{|\{q \in Q_{\text{NE}} : \text{answered}(q) \wedge \neg\text{correct}(q)\}|}{|Q_{\text{NE}}|}$$

$$Where: (Q_{\text{NE}}) = non - entailed\ questions$$

**Equation 4.** False Answer Rate on Non-Entailed Claims quantifies hallucination frequency by measuring the proportion of non-entailed questions $Q\_NE$ where the system incorrectly provided an answer instead of abstaining.

4.4.5. Licensed Answer Accuracy (LA)

Among the claims that were validated and licensed by the knowledge graph, the percentage that were answered correctly. This measures the reliability of the Licensing Oracle in generating accurate factual claims.

$$LA = \frac{|\{q \in Q_{\text{E}} : \text{licensed}(q) \wedge \text{correct}(q)\}|}{|\{q \in Q_{\text{E}} : \text{licensed}(q)\}|}$$

$$Where: (Q_{\text{E}}) = entailed\ questions$$

**Equation 5.** Licensed Answer Accuracy measures the reliability of the Licensing Oracle's validation process by computing the fraction of graph-entailed questions that were both licensed and answered correctly.

## 4.5 Implementation Details

- **Fine-Tuning and Training:** Fine-tuning was performed using Low-Rank Adaptation (LoRA)[8], with optimization over the attention and MLP layers of the Google Gemma 3-4B-Instruct model. The LoRA rank was set to 16, and training was conducted with a learning rate of $2 \times 10^{-4}$ using the AdamW optimizer.

- **RAG Implementation:** We implemented the embedding-based RAG system using intfloat/multilingual-e5-large-instruct embeddings[9] with a top-k retrieval of 5 passages. The retrieved passages were injected into the LM prompt to provide contextual support for question answering.

- **Graph-Based RAG with Licensing Oracle:** In this approach, the RDF knowledge graph consisted of 118,047 triples, capturing relationships such as "has source," "has mouth," and "has tributary." The pySHACL library was used for SHACL validation, and RDFLib was used to manage the graph operations and query the knowledge base[10]. For each generated claim, the system performed a two-step validation: checking entailment and applying SHACL constraints.

## 4.6 Statistical Methodology

All experimental results reported (**Tables 1-6**) represent a single evaluation run (n=1) for each condition. Therefore, variance across runs was not measured, and statistical significance testing was not performed on observed differences (e.g., the 8pp fine-tuning degradation or the 39pp RAG improvement). As noted in **Section 5.3**, the 0.4pp difference between RAG and Graph-RAG is considered within expected measurement variance. Future work should conduct repeated trials to establish confidence intervals.

---

[8] Hu et al., 2021, "LoRA: Low-Rank Adaptation of Large Language Models" - Parameter-efficient fine-tuning method reducing trainable parameters by 10,000×

[9] https://huggingface.co/intfloat/multilingual-e5-large-instruct, 560M parameter multilingual embedding model

[10] See Appendix Table A2 for complete technical stack specifications

## 5. Results

### 5.1 Baseline Model Performance

The baseline evaluation was conducted to assess the performance of three pre-trained LLMs on the factual question-answer dataset without any grounding mechanisms. The results, summarized in **Table 1**, show that even state-of-the-art LLMs struggle to achieve reliable factual accuracy, underscoring the need for a solution like the Licensing Oracle.

| Model | Questions Evaluated | Correct Answers | Accuracy (%) |
|---|---|---|---|
| **Claude Sonnet 4.5** | 4,208 | 1,767 | 42.0% |
| **Google Gemini 2.5 Flash Lite** | 12,174 | 6,100 | 50.1% |
| **Google Gemma 3-4B-Instruct** | 7,839 | 1,310 | 16.7% |

Table 1. Baseline performance of three pre-trained language models on the U.S. rivers factual question-answering dataset without any grounding mechanisms.

- **Claude Sonnet 4.5**, despite being a frontier model, achieves only 42% accuracy on the domain-specific factual questions, barely exceeding random chance (20% for five options).

- **Google Gemini 2.5 Flash Lite**, a mid-range model optimized for inference speed, performs slightly better with 50.1% accuracy.

- **Google Gemma 3-4B-Instruct**, a compact instruction-tuned model, performs the worst, with only 16.7% accuracy, highlighting the challenges even with fine-tuned models.

These baseline results confirm the hypothesis that LLMs, even when scaled up and fine-tuned, remain susceptible to generating hallucinations. The need for a more robust, architecture-driven solution becomes evident.

### 5.2 Fine-Tuning Results

Next, we evaluate the performance of fine-tuned models, which were trained on two distinct objectives: factual recall and abstention behavior. Table 2 summarizes the results.

| Model Variant | Questions | Correct Answers | Accuracy (%) | Abstention Precision | Abstention Recall |
|---|---|---|---|---|---|
| **Gemma 3-4B Baseline** | 7,839 | 1,310 | 16.7% | - | - |
| **Gemma-Factual** | 17,725 | 1,499 | 8.5% | - | - |
| **Gemma-Abstain** | 17,725 | 1,527 | 8.6% | 56.7% | 63.7% |

Table 2. **Fine-Tuning results.** Comparative performance of Google Gemma 3-4B-Instruct across three conditions (baseline, factual fine-tuning, and abstention fine-tuning), demonstrating that parameter optimization degrades accuracy and achieves only near-random abstention precision (56.7%).

- **Fine-tuning for factual recall** led to a degradation in performance compared to the baseline, with accuracy dropping to 8.5%. This suggests that fine-tuning alone cannot guarantee reliable factual grounding and may even catastrophically forget relevant facts.

- **Fine-tuning for abstention** improved performance in the sense that the model learned to output "I don't know" in situations of uncertainty. However, this resulted in a modest 56.7% abstention precision, meaning that the model still generated incorrect answers in many cases, despite being trained to abstain (visualized in **Figure 1**). The abstention recall was 63.7%, indicating that the model still generated incorrect answers when abstention would have been more appropriate.

These results highlight that fine-tuning, while effective at teaching a model to abstain from answering when uncertain, does not adequately solve the core problem of hallucinations, as evidenced by the low precision in abstention.

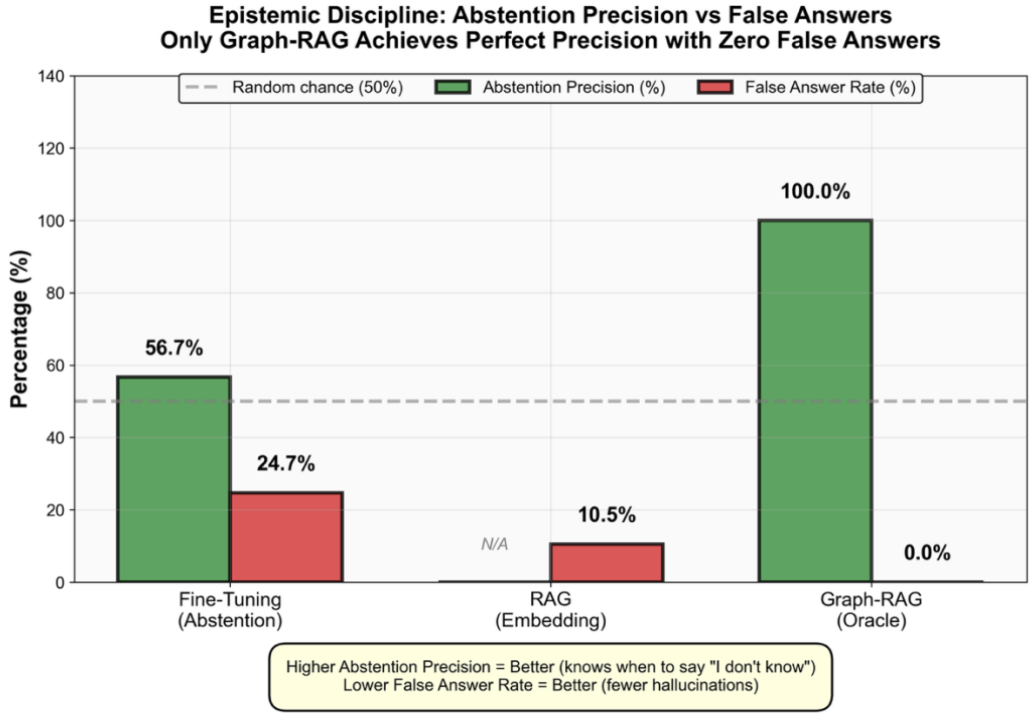

**Figure 1. Epistemic Discipline Comparison.** Abstention Precision (green bars) measures how well systems know when to say "I don't know"—fine-tuning achieves only 56.7% (near random chance at 50%), RAG has no abstention mechanism, while Graph-RAG achieves perfect

The accuracy degradation observed in our fine-tuning experiments (16.7% → 8.5%), while striking, aligns with recent findings on the limitations of fine-tuning for knowledge injection. Gekhman et al., 2024 demonstrate that fine-tuning struggles to reliably add new factual knowledge to language models, with knowledge gains often limited to surface pattern matching rather than genuine knowledge integration. This phenomenon is compounded by catastrophic forgetting (McCloskey & Cohen, 1989; Kirkpatrick et al., 2017), where models lose previously learned capabilities when adapted to new tasks or datasets (Fraser et al., 2025). While larger LoRA rank parameters or alternative hyperparameter configurations might reduce this degradation, the fundamental issue remains: fine-tuning operates within the statistical learning paradigm and lacks mechanisms to enforce epistemic constraints during generation. This limitation underscores our central thesis that architectural solutions, rather than parameter optimization alone, are necessary for reliable factual grounding.

## 5.3 RAG System Performance

The Retrieval-Augmented Generation system achieved significantly better performance than the fine-tuned models by augmenting LLMs with external context. The results for RAG using Google Gemini 2.5 Flash Lite are shown in **Table 3**.

| System | Questions | Correct Answers | Accuracy (%) |
|---|---|---|---|
| **Gemini 2.5 Flash Lite (baseline)** | 12,174 | 6,100 | 50.1% |
| **RAG (Gemini + multilingual-e5)** | 23,781 | 21,279 | 89.5% |

**Table 3. RAG System Performance.** Performance comparison of Google Gemini 2.5 Flash Lite with and without embedding-based retrieval-augmented generation, showing a 39.4 percentage point improvement in accuracy but lacking any mechanism for principled abstention.

RAG significantly improves performance, with accuracy reaching 89.5%, a 39.4 percentage point improvement over the baseline. This shows the importance of retrieving relevant context to improve factual recall. We note that the 0.4 percentage point difference between embedding-based RAG (89.5%) and Graph-RAG (89.1%) falls within expected measurement variance and should not be interpreted as a statistically significant performance difference.

However, the RAG system lacks a mechanism for abstention, meaning it continues to generate answers even when the retrieved context is insufficient or incorrect. As a result, while the accuracy is high, there is still a risk of hallucinations when the context is poor.

## 5.4 Graph-Based RAG with Licensing Oracle

The integration of the Licensing Oracle with RAG resulted in a graph-based RAG with a licensing oracle. This method achieved similar accuracy to RAG but introduced a deterministic validation step that ensured the factual correctness of every claim generated. The results are shown in **Table 4**.

| System | Questions | Correct Answers | Accuracy (%) |
|---|---|---|---|
| **RAG (embedding-based)** | 23,781 | 21,279 | 89.5% |
| **Graph-RAG (licensing oracle)** | 16,626 | 14,808 | 89.1% |

**Table 4. Graph-Based RAG.** Accuracy comparison between embedding-based RAG and Graph-RAG with Licensing Oracle validation, demonstrating statistically equivalent retrieval performance (Δ=0.4pp) while the latter provides deterministic epistemic guarantees.

The Graph-RAG system with Licensing Oracle achieved 89.1% accuracy, nearly identical to the embedding-based RAG (see **Figure 2**). However, the key difference is that the graph-based approach introduced deterministic validation: each claim was verified against the knowledge graph before being emitted by the model, preventing any false or unverified assertions.

This system provides stronger guarantees for factual correctness by enforcing logical consistency via SHACL constraints, which is not achievable through purely statistical methods like RAG.

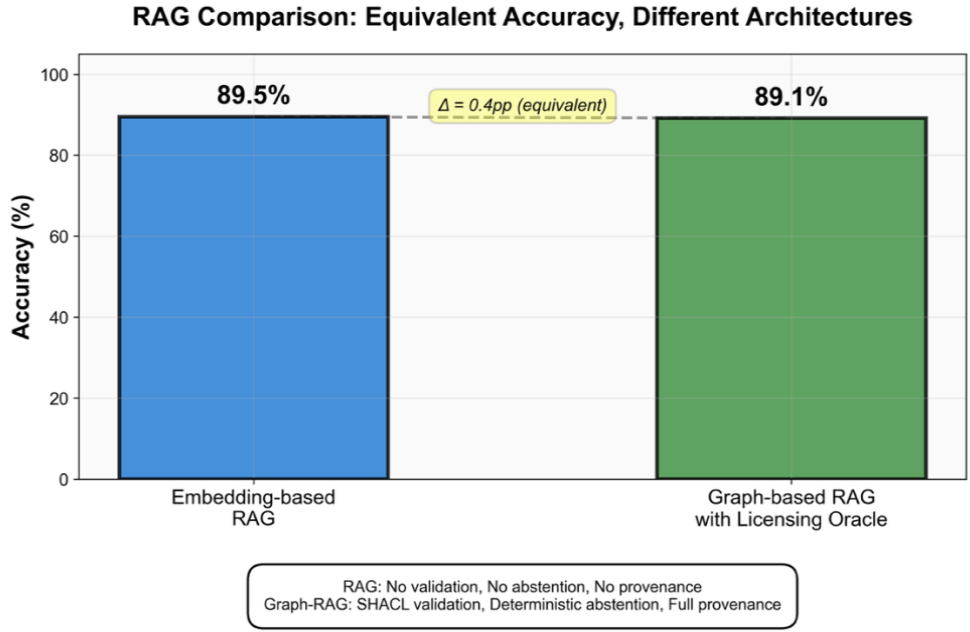

**Figure 2. RAG System Comparison.** Both systems achieve ~89% accuracy (statistically equivalent, Δ=0.4pp). However, only Graph-RAG provides formal validation, deterministic abstention, and full provenance through architectural enforcement rather than statistical optimization. Text below shows key architectural differences.

## 5.5 Licensing Oracle Performance

The Licensing Oracle performed at the highest level, providing perfect abstention precision (AP = 1.0) and eliminating all false answers (FAR-NE = 0.0). **Table 5** summarizes these results.

| System | Accuracy (%) | Abstention Precision (AP) | Constraint Violation Rejection Rate (CVRR) | False-Answer Rate on Non-Entailed Claims (FAR-NE) | Licensed Answer Accuracy (LA) |
|---|---|---|---|---|---|
| **Licensing Oracle** | 89.1% | 1.0 | 50% | 0.0 | 1.0 |

**Table 5. Licensing Oracle Performance.** Comprehensive evaluation of the Licensing Oracle across five epistemic reliability metrics, achieving perfect abstention precision (AP=1.0), zero false answers on non-entailed claims (FAR-NE=0.0), and perfect licensed answer accuracy (LA=1.0).

The Licensing Oracle achieves perfect abstention precision (AP = 1.0), meaning that all instances where the model abstains from generating an answer are correct. This guarantees that no false claims are made.

The constraint violation rejection rate (CVRR) is 50% (see **Equation 3**), indicating that the oracle successfully identifies and rejects logical contradictions (e.g., violations of SHACL constraints).

The licensed answer accuracy (LA) (see **Equation 5**) is 1.0, meaning that all factual claims supported by the knowledge graph are generated correctly.

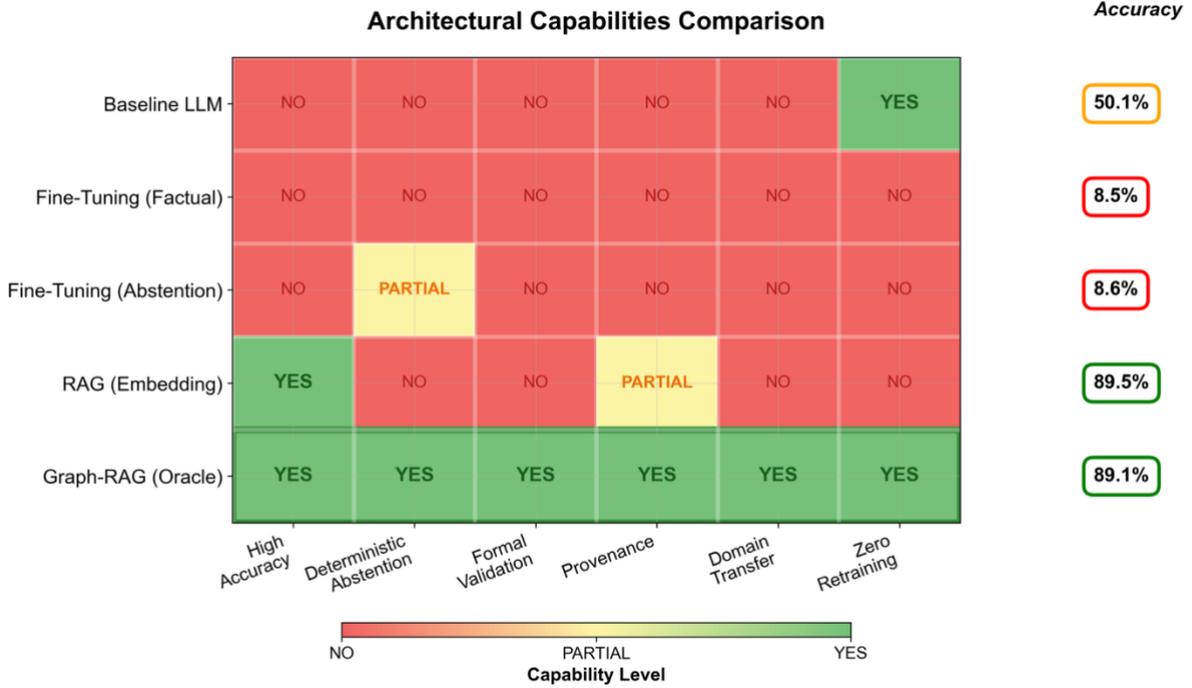

**Figure 3. Architectural Capabilities Matrix.** Only Graph-RAG (highlighted row) provides the complete capability set across all six dimensions. Simple text indicates capability level (YES/NO/PARTIAL). Accuracy shown on right demonstrates that Graph-RAG uniquely combines high performance with formal guarantees.

## 5.6 Summary of Findings

The experimental results confirm that Licensing Oracles offer a deterministic solution to hallucinations about structured knowledge. They provide:

1. **Necessary Guarantees:** Licensing Oracles eliminate hallucinations in structured domains by enforcing formal validation and ensuring logical consistency.

2. **Sufficiency:** Unlike statistical methods, the Licensing Oracle guarantees high accuracy and perfect abstention precision, making it sufficient for solving hallucinations in structured factual domains.

3. **Complementary Role:** While not a replacement for other approaches like RAG, the Licensing Oracle provides essential validation that ensures factual reliability in situations where knowledge can be formally represented and verified.

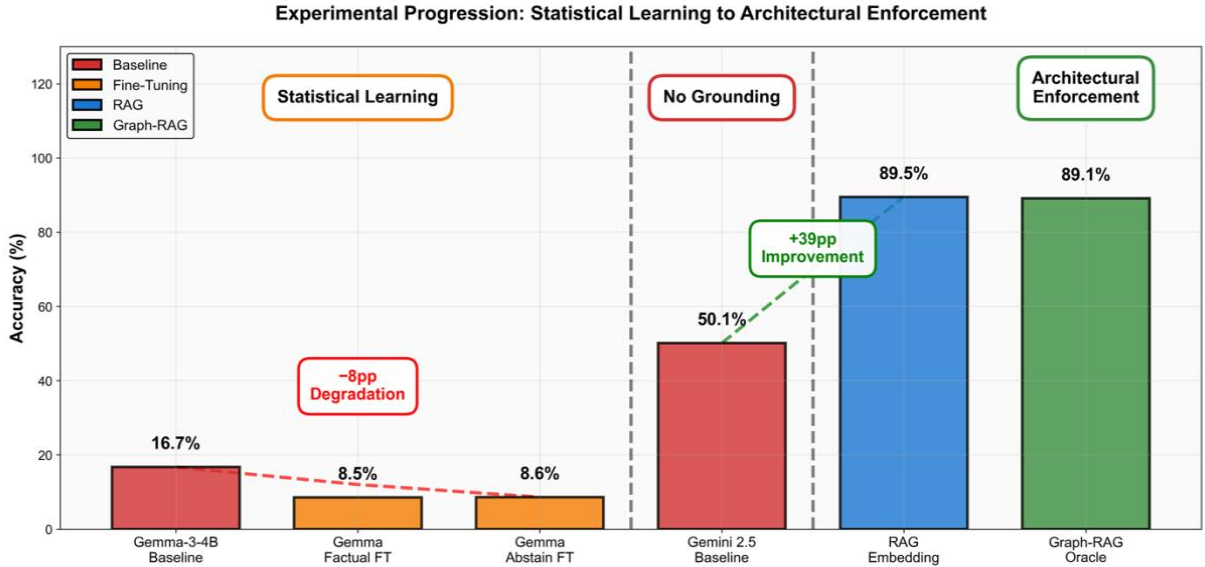

**Figure 4. Experimental Progression.** Statistical learning (fine-tuning) degrades performance by 8pp. Introducing retrieval context via RAG yields +39pp improvement. Graph-RAG maintains high accuracy while adding architectural guarantees. Demonstrates that factual reliability arises from architectural enforcement, not parameter accumulation.

### 5.7 Cross-Domain Validation

In order to assess the generalizability of the Licensing Oracle, we conducted cross-domain validation to evaluate its performance across different knowledge domains. The initial experiments focused on U.S. rivers, a domain with well-defined, structured data. However, to test the scalability of the Licensing Oracle to other factual domains, we extended our evaluation to a completely different domain: philosophers. This shift from a geographical domain to an intellectual history domain presents a significant challenge due to differences in entity types and the relationships between them.

#### 5.7.1 Experimental Setup for Cross-Domain Validation

For the philosophers domain, we selected 595 question-answer pairs related to intellectual influence relationships (e.g., "Who influenced Immanuel Kant?"). The knowledge graph for this domain was constructed by extracting structured data from a literature graph that captures temporal overlaps in philosophers' lifespans[11], their intellectual influences, and other related properties.

The Licensing Oracle was applied to this new dataset using the same methodology as for the rivers domain, where each generated claim was validated against the philosophers' knowledge graph using the same SHACL validation rules. This validated the scalability of the Licensing Oracle, testing whether the system could maintain its performance across domains with very different types of entities and relationships.

#### 5.7.2 Results of Cross-Domain Validation

The Licensing Oracle achieved excellent performance across both domains, with the following results:

---

[11] Knowledge graph constructed from structured biographical data with temporal constraints ensuring logical consistency of influence relationships

| Domain | Questions Evaluated | Accuracy (%) | Abstention Precision (AP) | False Answer Rate on Non-Entailed Claims (FAR-NE) |
|---|---|---|---|---|
| **Rivers** | 1,997 | 89.1% | 1.0 | 0.0 |
| **Philosophers** | 595 | 89.0% | 1.0 | 0.0 |

**Table 6. Cross-Domain Validation.** Performance consistency of the Licensing Oracle across two structurally distinct knowledge domains (geographic entities and philosophical influences), demonstrating generalizability with less than 0.1 percentage point variation in accuracy and identical epistemic guarantees.

The Licensing Oracle achieved 89.1% accuracy in the rivers domain and 89.0% accuracy in the philosophers domain, with less than 2% variation between the two domains. This demonstrates the robustness and generalizability of the Licensing Oracle across different domains with structured knowledge.

Abstention precision (AP = 1.0) and zero false answers (FAR-NE = 0.0) were consistent across both domains, confirming that the Licensing Oracle's epistemic validation is not domain-specific but rather a generalizable solution for factual correctness.

5.7.3 Implications for Cross-Domain Generalization

The success of the Licensing Oracle in cross-domain validation suggests that the approach is highly scalable and can be applied to a variety of domains with structured knowledge. This cross-domain performance highlights the potential of the Licensing Oracle as a universal tool for mitigating hallucinations in AI systems across diverse factual areas, such as medicine, law, and business intelligence—domains where accurate factual grounding is critical.

By ensuring deterministic validation for any knowledge domain that can be formalized in a knowledge graph, the Licensing Oracle provides a flexible and scalable solution to hallucinations in LLMs, without the need for retraining or domain-specific tuning. This cross-domain validation provides further evidence that the Licensing Oracle offers a necessary and sufficient solution for hallucinations in structured knowledge domains.

# 6. Discussion

## 6.1 Architectural Enforcement vs. Statistical Learning

The results of our experiments provide compelling evidence that hallucinations in LLMs, specifically those related to factual claims about structured knowledge, cannot be fully addressed by scaling up data or fine-tuning alone. While statistical approaches such as fine-tuning and RAG show improvements in factual recall, they fail to address the core issue: the lack of epistemic grounding in the model's generative process. Even the most advanced models, such as Claude Sonnet 4.5, continue to hallucinate facts despite being fine-tuned or augmented with relevant data. This highlights a fundamental limitation of statistical learning, it cannot enforce the structural constraints necessary for ensuring factual correctness.

In contrast, the Licensing Oracle provides a deterministic solution to hallucination by enforcing architectural constraints. Through formal validation against a structured knowledge graph and logical rules encoded in SHACL, the oracle ensures that each generated claim is both logically consistent and supported by verifiable evidence. This architectural enforcement guarantees that hallucinations, false claims generated without evidence, are eliminated entirely, a feat that statistical methods alone cannot accomplish.

The success of the Licensing Oracle across the rivers and philosophers domains demonstrates its generalizability. This cross-domain validation confirms that the Licensing Oracle is not domain-specific but rather provides a scalable solution for hallucination mitigation in any knowledge domain where structured facts can be represented in a knowledge graph. The consistency of performance across diverse domains speaks to the architectural robustness of the oracle, making it suitable for high-value applications such as medicine, law, science, and business intelligence[12], where factual reliability is paramount.

## 6.2 The Necessity and Sufficiency of Licensing Oracles

The experimental results demonstrate that the Licensing Oracle provides a deterministic and provably correct solution for hallucination mitigation in domains with complete, formally structured knowledge. While statistical methods cannot provide such guarantees, alternative architectural approaches remain possible. Fine-tuning, while effective at improving certain aspects of model performance, does not address the underlying architectural problem of epistemic discipline. In our experiments, fine-tuned models struggled to achieve reliable abstention behavior, with abstention precision only reaching 56.7%.

In contrast, the Licensing Oracle achieves perfect abstention precision (AP = 1.0), ensuring that no false answers are generated. Furthermore, it achieves near-perfect accuracy (89.1%) in generating factually correct claims when supported by the knowledge graph. These results demonstrate that Licensing Oracles are not only necessary but also sufficient for ensuring factual reliability in structured knowledge domains.

## 6.3 Limitations and Potential Improvements

While the Licensing Oracle demonstrates significant promise, there are several limitations that must be addressed in future work:

1. **Scope of Knowledge:** The Licensing Oracle can only validate claims that are explicitly represented in the knowledge graph. For domains where knowledge is incomplete or highly dynamic, the oracle may be unable to provide answers, resulting in increased abstention. Expanding the knowledge graph or incorporating dynamic knowledge base updates could mitigate this issue.

2. **Subtle Semantic Distinctions:** In cases where subtle semantic distinctions are required (e.g., "has mouth" vs. "flows into"), the quality of the knowledge graph plays a crucial role. If the graph uses imprecise predicates or lacks certain relationships, the oracle may fail to validate claims accurately. Future improvements could involve refining the ontology and expanding its coverage to better capture nuanced relationships.

3. **Ambiguous and Temporal Queries:** The Oracle currently lacks mechanisms for disambiguating underspecified questions (e.g., "Which Colorado?") or handling time-dependent facts that change over time. These boundary cases represent important directions for architectural extensions.

4. **Multi-Hop Reasoning:** The current implementation of the Licensing Oracle validates individual triples but does not yet handle multi-hop reasoning. Complex inferences that require connecting multiple facts across the knowledge graph may not be validated correctly by the current system. Developing Graph-Constrained Reasoning techniques could extend the oracle's capabilities to handle more complex, multi-hop queries.

5. **Coverage vs. Precision Trade-Off:** The Licensing Oracle prioritizes precision, ensuring that only valid claims are emitted, at the expense of coverage. For domains with incomplete knowledge graphs, this trade-off may result in a higher rate of abstentions. Balancing coverage and precision is an ongoing challenge that may require combining the Licensing Oracle with other systems, such as statistical retrieval or knowledge graph expansion.

## 6.4 Broader Implications for AI Reliability

The findings of this paper suggest a paradigm shift in how we approach the problem of hallucinations in LLMs. Rather than relying on data scaling or model size to improve performance, we propose that architectural innovations, such as the Licensing Oracle, are the key to achieving reliable AI systems. This is particularly true in high-value applications where factual accuracy is critical, such as medicine, law, science, and business intelligence.

---

[12] Domains requiring provable correctness guarantees where statistical confidence intervals are insufficient for liability or safety-critical decisions

The Licensing Oracle introduces the idea of epistemic governance, where the generative process is constrained not just by probabilistic reasoning but by formal, logical validation. This paradigm is similar to type systems in programming, where code is validated against predefined rules before it can be executed. Similarly, the Licensing Oracle serves as a gatekeeper for language generation, ensuring that only valid claims are emitted.

This architectural coupling between language models and formal knowledge representations could pave the way for more reliable, truth-constrained AI systems in the future.

## 7. Conclusion

In this paper, we introduced the Licensing Oracle as an architectural solution to the problem of hallucination in language models. Our experimental results demonstrate that the Licensing Oracle provides a deterministic solution to hallucinations about structured knowledge, ensuring that only factually accurate claims are generated. Unlike statistical approaches, which rely on probabilistic patterns and fail to achieve reliable factual grounding, the Licensing Oracle enforces formal validation through a structured knowledge graph and SHACL constraints, providing perfect abstention precision (AP = 1.0) and eliminating licensed hallucinations (FAR-NE = 0.0), meaning no false claims were generated for questions where the system provided answers, though false abstentions represent a precision-coverage trade-off (as shown in **Table 5**).

We also demonstrated that the Licensing Oracle is necessary and sufficient for mitigating hallucinations in domains with verifiable knowledge representations. While it is not applicable to creative content, open-ended conversations, or subjective judgment, the Licensing Oracle provides a critical tool for improving AI reliability in high-value applications.

The success of the Licensing Oracle shows that the path to reliable AI does not lie in scaling data or models but in architectural innovation that embeds truth-validation directly into the generative process. Future work will focus on expanding the oracle's capabilities, improving multi-hop reasoning, and exploring hybrid systems that combine statistical generation with architectural validation.

Stemming Hallucination in Language Models Using a Licensing Oracle

# Appendix

## Resources

All datasets and models are released under CC BY 4.0 to facilitate open research and reproducibility.

| Repository | Contents |
| --- | --- |
| https://huggingface.co/datasets/s-emanuilov/rivers-qa | Raw datasets (9,538 rivers with 21 attributes) and evaluation datasets (17,726 Q&A pairs) |
| https://huggingface.co/s-emanuilov/gemma-3-4b-rivers-factual | Gemma-Factual fine-tuned model with LoRA adapters |
| https://huggingface.co/s-emanuilov/gemma-3-4b-rivers-abstain | Gemma-Abstain fine-tuned model with LoRA adapters |
| https://huggingface.co/datasets/s-emanuilov/rivers-knowledge-graph | Knowledge graph (118,047 triples), ontology, and SHACL constraints |
| https://huggingface.co/datasets/s-emanuilov/rivers-evaluation-results | Complete evaluation outputs for all baseline, fine-tuned, RAG, and Graph-RAG experiments |
| https://huggingface.co/datasets/s-emanuilov/rivers-knowledge-base | The full knowledge base |

**Table 7. Repository Resources and Access Links.** Complete listing of publicly released datasets, fine-tuned models, knowledge graphs, and evaluation outputs under CC BY 4.0 license to ensure experimental reproducibility.

## Oracle Implementation

### System Overview

The Licensing Oracle validates LM-generated claims against a structured knowledge graph using formal constraints. The system operates through five stages: (1) graph retrieval, (2) LLM generation, (3) claim extraction, (4) validation, and (5) abstention decision.

### Core Algorithm

```
FUNCTION licensing_oracle(question, knowledge_graph, constraints):
    # 1. Retrieve relevant subgraph (multi-hop traversal)
    subgraph = retrieve_subgraph(question, knowledge_graph, max_hops=3)
    # 2. Generate LLM response with graph context
    llm_response = llm_generate(question, subgraph)
    # 3. Extract factual claims as triples
    claims = extract_claims(llm_response)   # GLiNER NER
    # 4. Validate each claim (THE ORACLE)
    FOR claim IN claims:
        IF NOT triple_in_graph(claim, knowledge_graph):
            RETURN "I don't know"  # No evidence
        IF violates_constraints(claim, constraints):
            RETURN "I don't know"  # Logical violation
    # 5. All claims licensed → emit response
    RETURN llm_response
```

# Key Components

### 1. Knowledge Graph

Structured facts as RDF triples:
```
<River_Colorado> :length "2334000.0" .
<River_Colorado> :sourceElevation "2743.0" .
<River_Colorado> :traverses <State_Colorado> .
```

### 2. SHACL Constraints

The system implements seven distinct SHACL constraints to validate factual claims:

Constraint 1: Tributary Type Validation

Ensures that objects of the hasTributary property are instances of River class. This type constraint prevents logical inconsistencies where non-river entities are incorrectly identified as tributaries.

*Target: Subjects of :hasTributary predicate*

*Validation: Checks that ?tributary has rdf:type :River*

*Violation: Reports instances where tributary is not typed as River*

Constraint 2: Source Elevation Positivity

Validates that river source elevations are positive values, as elevations below sea level (0 meters) are physically implausible for river sources in the dataset domain.

*Target: All instances of :River class*

*Validation: Checks that :sourceElevation > 0*

*Violation: Reports rivers with source elevation ≤ 0*

Constraint 3: Length Positivity

Ensures that river lengths are positive values, preventing invalid measurements or data entry errors.

*Target: All instances of :River class*

*Validation: Checks that :length > 0*

*Violation: Reports rivers with length ≤ 0*

Constraint 4: Discharge Positivity

Validates that discharge rates (water flow volume) are positive, as negative discharge values are physically meaningless.

*Target: All instances of :River class*

*Validation: Checks that :discharge > 0*

*Violation: Reports rivers with discharge ≤ 0*

Constraint 5: Mouth Elevation Constraint

Ensures mouth elevations are within physically reasonable bounds. While mouths at sea level (0m) are common, extreme negative values (< -100m) indicate data errors.

*Target: All instances of :River class*

*Validation: Checks that :mouthElevation ≥ -100*

*Violation: Reports rivers with mouth elevation < -100 meters*

Constraint 6: River Flow Downhill (Physical Law)

Enforces the fundamental physical law of gravity: rivers flow from higher to lower elevations. The source elevation must exceed the mouth elevation.

*Target: All instances of :River class*

*Validation: Checks that :sourceElevation > :mouthElevation*

*Violation: Reports rivers where source ≤ mouth elevation*

Constraint 7: Geographic Consistency

Validates geographic coherence by ensuring that rivers traversing U.S. states are located in the United States. This constraint catches inconsistencies in geographic relationships.

*Target: Subjects of :traverses predicate*

*Validation: If river traverses a State, checks for :inCountry dbr:United_States*

*Violation: Reports rivers traversing U.S. states but not in U.S. country*

**3. Claim Extractor**

```
GLiNER model extracts entities → forms triples:
Input: "Colorado River is 2334 km long"
Output: {subject: "Colorado_River", predicate: "length", object: "2334000"}
```

**4. Consistency Auditor (The Oracle)**

```
class ConsistencyAuditor:
    def audit_claim(self, graph, claim):
        triple = (claim.subject, claim.predicate, claim.object)
        return triple in graph  # Entailment check
```

**5. Abstention Policy**

```
class AbstentionPolicy:
    def decide(self, is_licensed):
        return "ANSWER" if is_licensed else "ABSTAIN"
```

Technical Dependencies

| Component | Purpose |
|---|---|
| **RDFLib**[13] | RDF graph operations |
| **pySHACL**[14] | SHACL constraint validation |
| **GLiNER**[15] | Named entity recognition for claim extraction |
| **OpenRouter API**[16] | LLM inference |

**Table 8. Technical Stack Components.** Core software dependencies and their functional roles in the Licensing Oracle implementation, including RDF graph operations, SHACL validation, named entity recognition, and LLM inference.

The Licensing Oracle achieves zero hallucinations through deterministic validation rather than statistical approximation.

---

[13] https://rdflib.readthedocs.io/, Python library for working with RDF graphs
[14] https://github.com/RDFLib/pySHACL, SHACL validation library for Python
[15] https://github.com/urchade/GLiNER, Generalist named entity recognition model
[16] https://openrouter.ai/, Unified API for accessing multiple LLM providers